# An Improved Heart Disease Prediction Using Stacked Ensemble Method

Md. Maidul Islam [1], Tanzina Nasrin Tania[1], Sharmin Akter[1], and Kazi Hassan Shakib[2]

[1] City University, Dhaka, Bangladesh
{rummanmaidul13, tanzinatania17, sharmin.cse051}@gmail.com
[2] Chittagong University of Engineering & Technology, Chittagong, Bangladesh
kazishakib98@gmail.com

**Abstract.** Heart disorder has just overtaken cancer as the world's biggest cause of mortality. Several cardiac failures, heart disease mortality, and diagnostic costs can all be reduced with early identification and treatment. Medical data is collected in large quantities by the healthcare industry, but it is not well mined. The discovery of previously unknown patterns and connections in this information can help with an improved decision when it comes to forecasting heart disorder risk. In the proposed study, we constructed an ML-based diagnostic system for heart illness forecasting, using a heart disorder dataset. We used data preprocessing techniques like outlier detection and removal, checking and removing missing entries, feature normalization, cross-validation, nine classification algorithms like RF, MLP, KNN, ETC, XGB, SVC, ADB, DT, and GBM, and eight classifier measuring performance metrics like ramification accuracy, precision, F1 score, specificity, ROC, sensitivity, log-loss, and Matthews' correlation coefficient, as well as eight classification performance evaluations. Our method can easily differentiate between people who have cardiac disease and those are normal. Receiver optimistic curves and also the region under the curves were determined by every classifier. Most of the classifiers, pretreatment strategies, validation methods, and performance assessment metrics for classification models have been discussed in this study. The performance of the proposed scheme has been confirmed, utilizing all of its capabilities. In this work, the impact of clinical decision support systems was evaluated using a stacked ensemble approach that included these nine algorithms.

**Keywords:** Prediction, Heart Disease, CART, GBM, Multilayer Perception.

## 1   Introduction

Heart disorder, which affects the heart and arteries, is one of the most devastating human diseases. The heart is unable to pump the required volume of blood toward other parts of the body when it suffers from cardiac problems. In the case of heart disease, the valves and heart muscles are particularly affected. Cardiac illness is also referred to as cardiovascular disease. The cardiovascular framework comprises all blood vessels, including arteries, veins, and capillaries, that constitute an intricate system of the bloodstream throughout the organ. Cardiovascular infections include cardiac illnesses, cerebrovascular infections, and artery illnesses. Heart disease may be a hazard, usually unavoidable and an imminent reason for casualty. Heart disease is currently a prominent issue with all other well-being ailments since many people are losing their lives due to heart disease. Cardiovascular disease kills 17.7 million people per year, accounting for 31% of all deaths globally, as per the World Health Organization (WHO). Heart attacks and strokes account for 85% of these cases. Heart-related disorders have also become the major cause of death in India [1]. In the United States, one person is killed every 34 seconds. [9]. Heart diseases killed 1.7 million Indians in 2016, concurring to the 2016 Worldwide Burden of Disease Report, released on September 15, 2017 [3]. According to a WHO report published in 2018, nearly 6 million people died globally in 2016 because of heart infections. [4]. Controlling heart disorders costs approximately 3% of total healthcare spending [20]. The World Health Organization's projections provided the impetus for this project. The WHO predicts that roughly 23.6 million people will die from heart disease by 2030. The expanding rate of heart infections has raised worldwide concern. Heart failure is tougher to diagnose because of diabetes, hypertension, hyperlipidemia, irregular ventricular rate, and other pertinent diagnosable conditions. As cardiac illness becomes increasingly common, data on the condition is getting more nonlinear, non-normal, association-structured, and complicated. As a result, forecasting heart illness is a major difficulty in medical data exploration, and clinicians find it extremely difficult to properly forecast heart disease diagnosis. Several studies have endeavored to use advanced approaches to analyze heart disease data. If the bagging is not adequately represented in the ensemble approach, it might result in excessive bias and consequently under-fitting. The boosting is also difficult to apply in real time due to the algorithm's increasing complexity. On the other hand, our proposed approach may combine the skills of several high-performing models on a classification or regression job to provide predictions that outperform any single model in the ensemble while also being simpler to build. Our suggested system hasn't received much attention; so we've attempted to build it correctly and come up with a nice outcome, and a superior prediction system.

The organization of the paper is explained as follows. In Section II, we have made an effort to state related research contributions, state their major contributions and compare with our work. We also provided a table with the underlying overview of the related works and comparison analytics for readers. With Section III, we have provided an outline of the system methodology and outlined the architecture. In section IV, implementations, and experimental results are described. Section V, we speak on our limitation in section and we conclude the paper

## 2      Literature Review

The study aims to look into how data mining techniques may be used to diagnose cardiac problems [15]. Practitioners and academics have previously employed pattern recognition and data mining technologies in the realm of diagnostics and healthcare for prediction purposes [13]. Various contributions have been made in recent times to determine the best preferred approach for predicting heart disorders [8]. So, the above part explores numerous analytical methodologies while providing a quick overview of the existing literature regarding heart disorders. In addition, current techniques have been evaluated in several ways, including a comprehensive comparison after this section.

Mohan S. et al. [1] developed a unique approach to determining which ML approaches are being used to increase the accuracy of heart illness forecasting. The forecast model is introduced using a variety of feature combinations and well-known classification methods. They attain an enhanced performance level with an accuracy level of 88.7% using the Hybrid Random Forest with Linear Model prediction model for heart disease (HRFLM). As previously stated, the (ML) techniques used in this study include DT, NB, DL, GLM, RF, LR, GBT, and SVM. All 13 characteristics as well as all ML techniques were used to reproduce the investigation.

Palaniappan S. et al. [2] applied a technology demonstrator Intelligent Heart Disease Prediction System (IHDPS), using data mining approaches such as DT, NB, and NN. The results show that each approach seems to have a different advantage in reaching the defined extraction criteria. Based on medical factors like sex, age, blood sugar, and blood pressure, this can forecast the probability of individuals developing heart disorders. It enables considerable knowledge to be established, such as patterns and correlations among galenic aspects connected to heart illness. The Microsoft.NET platform underpins IHDPS. The mining models are built by IHDPS using the CRISP-DM approach.

Bashir S. et al. [4] in their research study discusses how data science can be used to predict cardiac disease in the medical industry. Despite the fact that several studies have been undertaken on the issue, prediction accuracy still needs to be improved. As a result, the focus of this study is on attribute selection strategies as well as algorithms, with numerous heart disease datasets being utilized for testing and improving accuracy. Attribute selection methodologies such as DT, LR, Logistic Regression SVM, NB, and RF are used with the Rapid miner, and the results indicate an increase in efficiency.

Le, H.M. et al. [5] rank and weights of the Infinite Latent Feature Selection (ILFS) approach are used to weight and reorder HD characteristics in our method. A pulpous margin linear SVM is used to classify a subset of supplied qualities into discrete HD classes. The experiment makes use of the UCI Machine Learning Repository for Heart Disorders' universal dataset. Experiments revealed that it suggested a method is useful for making precise HD predictions; our tactic performed the best, with an accuracy of 90.65% as well as an AUC of 0.96 for discriminating 'No existence' HD from 'Existence' HD.

Yadav, D.C. and Pal et al. [6] implemented M5P, random Tree, and Reduced Error Pruning using the Random Forest Ensemble Method were presented and investigated as tree-based classification algorithms. All prediction-based methods were used after identifying features for the cardiac patient dataset. Three feature-based techniques were employed in this paper: PC, RFE, and LR. For improved prediction, the set of variables was evaluated using various feature selection approaches. With the findings, they concluded that the attribute selection methods PC and LR, along with the random-forest-ensemble approach, deliver 99% accuracy.

Kabir, P.B. and Akter, S. et al. [7] among the most fundamental and widely used ensemble learning algorithms are tree-based techniques. Tree-based models such as Random Forest (RF), and Decision Tree (DT), according to the study, provide valuable intelligence with enhanced efficiency, consistency, as well as application. Using the Feature Selection (FS) method, relevant features are discovered, and classifier output is produced using these features. FS eliminates non-essential characteristics without affecting learning outcomes. Our study aims to boost the performance. The aim of the research is really to apply FS in conjunction with tree-based approaches to increase heart disease prediction accuracy.

Islam, M.T. et al, [8] in this work, PCA has been used to decrease characteristics. Aside from the final clustering, a HGA with k-means was applied. For clustering data, the k-means approach is often applied. Because this is a heuristic approach, it is possible for it to become trapped in local optima. To avoid this problem, they used the HGA for data clustering. The suggested methodology has a prediction accuracy of 94.06 percent for early cardiac disease.

Rahman, M.J.U. et al, [10] the main purpose of this work is just to create a Robust Intelligent Heart Disease Prediction System (RIHDPS) applying several classifiers such as NB, LR, and NN. This content investigated the effectiveness of medical decision assistance systems utilizing ensemble techniques of these three algorithms. The fundamental purpose of this study is to establish a Robust Intelligent Heart Disease Prediction System (RIHDPS) by combining 3 data mining modelling techniques into an ensemble method: NB, LR, and NN.

Patel, J. et al, [12] utilizing W-E-K-A, this study evaluates alternative Decision Tree classification algorithms to improve contribution in heart disorder detection. The methods being tested include the J48 approach, the LMT approach, and the RF method. Using existing datasets of heart disease patients as from the UCI repository's Cleveland database, the performance of decision tree algorithms is examined and validated. The aim of the research is to utilize data mining tools that uncover hidden patterns in cases of heart problems as well as to forecast the existence of heart disorders in individuals, ranging from no existence to likely existence.

Bhatla, N. et al. [28] research aims to look at different data mining techniques that might be employed in computerized heart disorder forecasting systems. The NN with 15 features has the best accuracy (100%) so far, according to the data. DT, on either hand, looked impressive with 99.62 percent accuracy when using 15 characteristics. Furthermore, the Decision Tree has shown 99.2% efficiency when combined with the Genetic Algorithm and 6 characteristics.

**Table 1.** A literature evaluation of cardiac disease predictions included a comparison of several methods.

| Source | Datasets | FS | Attributes | Classifier & Validation techniques | Accuracy |
|---|---|---|---|---|---|

5| | | | | | |
|---|---|---|---|---|---|
| Mohan S. [1] | Cleveland UCI repository | HRFLM | 14 attributes | DT, GLM, RF, and 5 more attributes. | 88.4% |
| Bashir S. [4] | UCI dataset | Minimum Redundancy Maximum Relevance (MRMR) | FBS, Cp, Trestbps, Chol, Age, Slope, Sex, and more 7 attributes | NB, Logistic Regression, LR SVM, DT and RF | NB: 84.24% LR (SVM): 84.85% |
| Le, H.M. [5] | UCI Machine Learning Repository | Infinite Latent Feature Selection (ILFS) | 58 attributes | WEKE, NB, LR, Non-linear SVM (Gaussian, Polynomial, Sigmoid), and Linear SVM | Linear SVM: 89.93%, ILFS: 90.65% |
| Yadav D.C. and Pal [6] | UCI repository | Lasso Regularization, Recursive Features Elimination and Pearson Correlation | Resting, FBS, CP, Chol, Sex, Ca, Age, and 7 more attributes | Random Tree, M5P, and Reduced Error Pruning with Random Forest Ensemble Method | Random Forest ensemble method: 99% |
| Kabir P.B. and Akter S. [7] | Hungary (HU), Long Beach (LB), Cleveland (Cleve.), and Switzerland (SR) | Hybrid | Cordocentesis, Max HR achieved, Epoch, Triglyceride, Sign, Coronary Infarction, Diastolic Pressure, and 6 more attributes | LGBM, RF, NB, SVM, and 3 more algorithms | KNN: 100.00% DT: 100.00% RF: 100.00% |
| Islam M.T. [8] | UCI Machine Learning Repository | PCA | 14 attributes | H-G-A with k-means | 94.06% |
| Patel J. [12] | Cleveland UCI repository | WEKA | 13 attributes | DT(J48), LMT, RF | J48 tree technique: 56.76% |

## 3   Methodology

This section mentioned above proposes an advanced and efficient prediction of heart disease based on past historical training data. The ideal strategy is to analyze and test various data-mining algorithms and to implement the algorithm that gives out the highest accuracy. This research also consists of a visualization module in which the heart disease datasets are displayed in a diagrammatic representation using different data visualization techniques for user convenience and better understanding. The subsections that follow go through several materials and methodologies in detail. The research design is shown in Section A, the data collection and preprocessing are summarized in Section B, and the ML classification techniques and stacked ensemble approach are explained in Section C of this study.

### 3.1 Research Design

In this section, gather all of the data into a single dataset. This approach for extracting functions for cardiovascular disease prognostication may also be applied with this aspect analysis procedure. Following the identification of accessible data resources, those are additionally picked, cleansed, and then converted to the required distribution. The atypical identification survey provides valuable characteristics for predicting coronary artery disease prognosis. Cross-validation, several classification approaches, and the stacked ensemble method will be utilized to predict using pre-processed data. After completing all of these steps, the illness will be forecast favorably. Following that, we'll assess the entire performance. The outcome will be determined after the performance review.

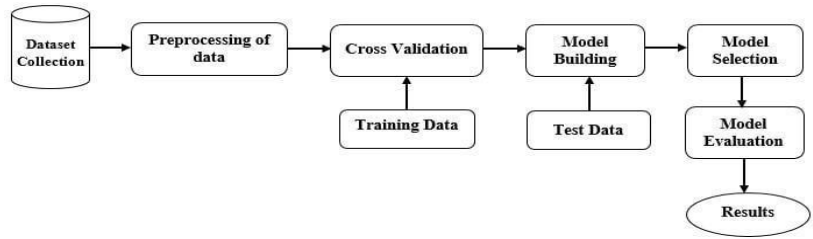

**Fig. 1.** Methodological framework of heart disease.

### 3.2 Data Collection & Preprocessing

In this study, we used Statlog, Cleveland, and Hungary datasets as the three datasets in this fact compilation. There are 1190 records in all, with 11 characteristics and one target variable. Chest pain, cholesterol, sex, resting blood pressure, age, resting ecg-normal (0), st-t abnormality (1), lv hypertrophy (2), fasting blood sugar max hate rate, exercise angina, old-peak, st slope-normal (0), upsloping (1), flat (2), downsloping(3), 0 denoting no disease and 1 denoting illness. It should be noted that null or missing values are utilized to represent zero values. As a result, we must delete null values throughout the data preparation step. But in our case, we have no null values. After that, we complete exploratory data analysis.

**Table 2.** Features of the dataset descriptive information.

| Features | Definition | Type |
|---|---|---|
| Age | Patient's age in years successfully completed | Numerical |

| Sex | Male patients are indicated at 1 and female patients are indicated at 0. | Nominal |
|---|---|---|
| Chest Pain | The four types of chest pain that patients feel are: 1. typical angina, 2. atypical angina, 3. non-anginal pain, and 4. asymptomatic angina. | Nominal |
| Resting BPS | Blood pressure in mm/HG while in resting mode | Numerical |
| Cholesterol | mg/dl cholesterol in the bloodstream | Numerical |
| Fasting Blood Sugar | Fasting blood sugar levels > 120 mg/dl are expressed as 1 in real cases and 0 in false cases. | Nominal |
| Resting ECG | The ECG results when at rest are displayed in three different values. 0: Normal 1: ST-T wave abnormality 2: Left ventricular hypertrophy. | Nominal |
| Max heart rate | Accomplished maximum heart rate | Numerical |
| Exercise angina | Exercise-induced angina 0 represents No and 1 represents Yes. | Nominal |
| Oldpeak | In compared to the resting state, exercise caused ST-depression. | Numerical |
| ST slope | Three values for the ST segment assessed of slope at peak exercise: 1. slanting, 2. flat, 3. slanting | Nominal |
| Target | It is the objective variable that we must forecast. A score of 1 indicates that the person is at risk for heart disease, whereas a value of 0 indicates that the person is in good health. | Numerical |

### 3.3 Models

Machine learning classification methods are utilized in this phase to classify cardiac patients and healthy people. The system employs RF Classifier, MLP, KNN, ET Classifier, XGBoost, SVC, AdaBoost Classifier, CART, and GBM, among other common classification techniques. For our suggested system, we will apply the stacked ensemble approach. We need to construct a base model and a meta-learner algorithm for a stacked ensemble. The most relevant and standard evaluation metrics for this problem area, such as sensitivity, specificity, precision, F1-Score, ROC, Log Loss and Mathew correlation coefficient are used to assess the outcome of each event.

1. RF Classifier: Random Forest Model is a classification technique that uses a random forest as its foundation. As in regression and classification, an algorithm may handle data sets with both continuous and categorical variables. It outperforms the competition when it comes to categorized problems. Criterion: this is a function that determines whether or not the split is correct. We utilized "entropy" for information gain, and "gini" stands for Gini impurity.

$$Gini = 1 - \sum_{i=1}^{G} (p_i)^2$$

$$Entropy = \sum_{i=1}^{G} - p_i * log_2 (p_i)$$

2. MLP: A pelleting neural network called a multi-layer perceptron (MLP) establishes a number of outputs from a collection of inputs. Multiple sections of input nodes comprise an MLP, between the inlet and outlet layers is linked as a directed graph.
3. KNN: K-NN method is straightforward to implement and does not require the use of a hypothesis or any other constraints. This algorithm may be used to do exploration, validation, and categorization. Despite the fact that K-NN is the most straightforward approach, it is hampered by duplicated and unnecessary data.
4. Extra Tree Classifier: Extremely Randomized Trees, or Extra Trees, is a machine learning ensemble technique. This is a decision tree ensemble comparable like bootstrap aggregation and random forest, among other decision tree ensemble, approaches. The Extra Trees approach uses the training data to construct a significant number of extremely randomized decision trees. An Average of decision tree estimates is used in regression, whereas a democratic majority is utilized in classification.
5. XGBoost: The XGBoost classifier is a machine learning method for categorizing both structured and tabular data. XGBoost is a high-speed and high-performance gradient boosted decision tree implementation. XGBoost is a high-gradient gradient boost algorithm. As a result, it's a complicated machine learning method with many moving parts. XGBoost can handle large, complicated datasets with ease. XGBoost is an ensemble modelling approach.
6. SVC: In both classification and regression issues, the Support Vector Classifier (SVC) is a common supervised learning technique. The SVC method's purpose is to find the optimal path or set point for categorizing n-dimensional regions because the following observations may be readily classified. SVC can be used to select the extreme positions that aid in the construction of the hyperplane. The Support Vector Machine is the method, and support vector classifiers are prominent examples.
7. AdaBoost Classifier: The Algorithms, shorthand for Adaptive Boosting, is a boosting approach used in Machine Learning as Ensemble Learning. Each instance's weights are reassigned, with larger weights applied for instances that were mistakenly identified. This is known as "Adaptive Boosting".
8. CART: Data is divided up frequently based on a parameter in decision trees, a kind of supervised machine learning. In the training data, specify the input and the associated output. Two entities may be used to explain the tree: decision nodes and leaves.
9. GBM: Gradient boosting is a collection of classification algorithms that may be applied to a variety of issues such as classification and regression problems. It as-sembles a prediction system from a collection of weak frameworks, — usually decision trees.
10. Stacked Ensemble: The term "ensemble" relates to the procedure of combining many models. As a result, instead of employing model to make predictions, a group of models is used. Ensemble uses two different techniques:

    o Bagging creates a unique training segment with replenishment from experimental training phase, as well as the outcome is determined by a majority vote. Consider the Random Forest example.
    o Boosting transforms weak learners to strong learners through creating pursuant models with overall performance as the final model. For instance, in AdaBoost and XG BOOST.

The stacked ensemble approach will be used. The stacked ensemble approach would be a supervised ensemble classification strategy that stacks many prediction algorithms to find the optimum combination. Stacking, also called as Superior Training or Stacking Regression, is a set of computational where another second-level regression model "metalearner" is combined with a first-level regression model has been programmed to find the optimum possible combination of basic learners. Stacking, in contrast to bagging and boosting, aims to bring together strong, varied groups of learners. We have completed our work in the following sections

1. For this system, we import all of the necessary libraries.
2. After loading our dataset, we clean and preprocess it.
3. We use the z-score to identify and eliminate outliers.
4. We divided the data into two parts: training and testing, with 80/20 percentages.
5. We developed a model using cross-validation.
6. For a stacked ensemble technique, we stack all of the models such as RF, MLP, KNN, ETC, XGB, SVC, ADB, CART, and GBM.
7. We assess and compare our model to other models.

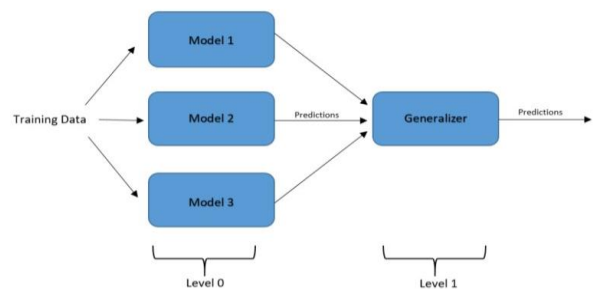

**Fig. 2.** Stacked Ensemble Method

Figure 2 depicts two levels: LEVEL 0 and LEVEL 1. First, we use the base learners (level 0) to make forecasts. The ensemble prediction is then generated by feeding those forecasts into the meta-learner (level 1).

## 4    Result Analysis

This section presents the outcomes of changing the ten orders indicated above. PRC, Sensitivity, Specificity, F1 Score, ROC, Log Loss, and MCC are the most common evaluation metrics used in this analysis. Complexity refers to a calculation that defines the importance of a segment of the review, whereas recall refers to the number of times genuinely qualified people are recovered.

**Table 3.** Result of various models with proposed model.

| Model | Accuracy | PRC | Sensitivity | Specificity | F1 Score | ROC | Log_Loss | MCC |
|---|---|---|---|---|---|---|---|---|
| Stacked Classifier | 0.910638 | 0.898438 | 0.934959 | 0.883929 | 0.916335 | 0.909444 | 3.086488 | 0.821276 |
| RF | 0.893617 | 0.865672 | 0.943089 | 0.839286 | 0.902724 | 0.891188 | 3.674399 | 0.789339 |
| MLP | 0.821277 | 0.809160 | 0.861789 | 0.776786 | 0.834646 | 0.819287 | 6.172973 | 0.642127 |
| KNN | 0.800000 | 0.787879 | 0.845528 | 0.750000 | 0.815686 | 0.797764 | 6.907851 | 0.599458 |
| Extra Tree Classifier | 0.885106 | 0.869231 | 0.918699 | 0.848214 | 0.893281 | 0.883457 | 3.968343 | 0.770445 |
| XGB | 0.897872 | 0.896000 | 0.910569 | 0.883929 | 0.903226 | 0.897249 | 3.527409 | 0.795248 |
| SVC | 0.812766 | 0.788321 | 0.878049 | 0.741071 | 0.830769 | 0.809560 | 6.466933 | 0.627138 |
| AdaBoost | 0.817021 | 0.812500 | 0.845528 | 0.785714 | 0.828685 | 0.815621 | 6.319943 | 0.633084 |
| CART | 0.851064 | 0.879310 | 0.829268 | 0.875000 | 0.853556 | 0.852134 | 5.144121 | 0.703554 |
| GBM | 0.829787 | 0.826772 | 0.853659 | 0.803571 | 0.840000 | 0.828615 | 5.879016 | 0.658666 |

The Stacked Ensemble Classifier, with an accuracy of 0.910, sensitivity of 0.934, specificity of 0.883, best f1-score of 0.916, minimum Log Loss of 3.08, and highest ROC value of 0.909, is the best performer. Of the same evaluation metrics in every region, Random Forest has the highest sensitivity level, while XGboost is second best.

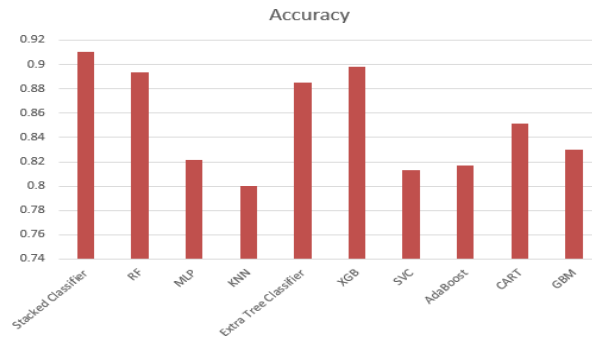

**Fig. 3.** Accuracy Chart of ML Models

This Figure-3 shows a visual depiction of effectiveness for all the other previously discussed machine learning techniques.

Stacked classifier model's accuracy is 91.06%, however, the F1 score is 0.9163. The accuracy of the XGB and RF algorithms, on the other hand, is 89.78% and 89.36%, respectively, with F1 scores of 0.8972 and 0.8911. The accuracies of Extra Tree Classifiers, CART, GBM, MLP, SVC, and KNN algorithms are 88.51%, 85.10%, 82.97%, 82.12%, 81.27%, and 80.00%.

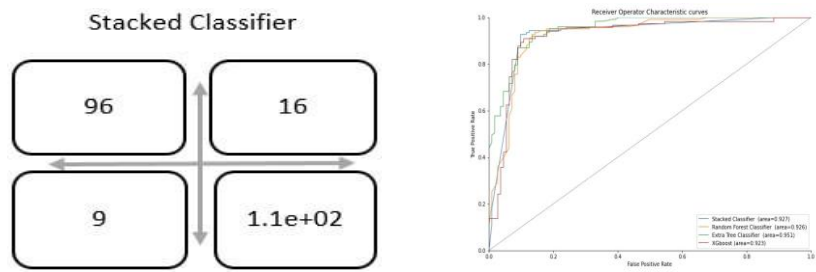

**Fig. 4.** Confusion Matrixes of Stacked Classifier Models and ROC Curve.

The confusion matrix for the implemented system is generated as shown in the diagram above. In the area of machine learning, extracted features are also referred to as artificial neurons. It is a statistical form that allows the reproduction of the results of an approach. In the case of graph partitioning, an ensemble learning approach is extremely useful. Knowledge is, specifically, the complexity of quantitative categorization.

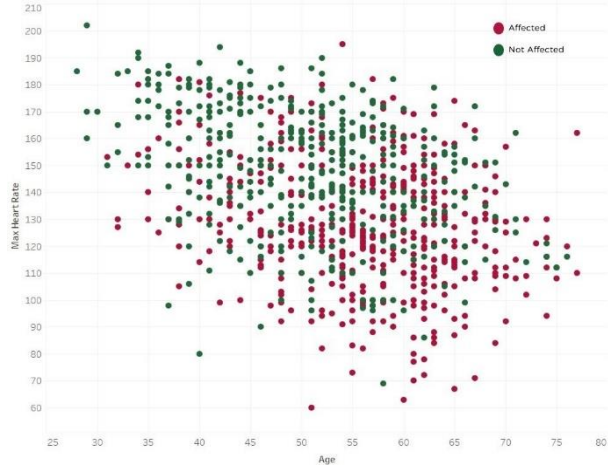

**Fig. 5.** Heart Disease Identification.

Figure-5 depicts a visual representation of all cardiac problems being detected. Crimson indicates a heart attack, whereas verdant indicates no cardiac disease.

## 5   Conclusion & Future Recommendation

Among the most significant threats to human survival is heart disease. Predicting cardiac illness has become a major concern and priority in the medical industry. Using the Stacked Ensemble Classifier, we have shown an improved heart disease prediction method. It incorporates a number of different prediction techniques. In this work, we examined the significance of prediction performance, precision, ROC sensitivity, Specificity, F1 Score, Log Loss, and MCC. To identify whether or not a person has a heart problem, we applied machine learning techniques. The medical data set was used in a variety of ways. As a consequence of the findings, we discovered that the enhanced stacked ensemble approach provides better accuracy than previous methods. The purpose of this research is to inquire about particular ML techniques on a form, therefore we further wanted to increase the dependability of the system's operations to provide a much adequate assertion as well as encourage certain Approaches for recognizing the appearance of CVD. The above-mentioned structure could be adapted and repurposed for new purposes. The results show that these data mining algorithms may accurately predict cardiac disease with a 91.06 percent accuracy rate. As our study is based on recorded data from the Statlog, Cleveland, and Hungary datasets, for future research possibilities, we will aim to train and test on a large medical data set using many ensemble methods in the future to see if we can enhance their performance. Our ensemble method is superior to traditional methods, as even if it is overfitting at times, it usually reduces variances, as well as minimizes modeling method bias. It also has superior Predictive performance, reduces dispersion and our approach has superior efficiency by choosing the best combination of models.